\documentclass[11pt]{article}

\usepackage[preprint]{acl}

\usepackage{times}
\usepackage{latexsym}

\usepackage[T1]{fontenc}
\usepackage[utf8]{inputenc}

\usepackage{microtype}

\usepackage{inconsolata}

\usepackage{graphicx}
\usepackage{amsmath}
\usepackage{booktabs}
\usepackage{tabularx}
\usepackage{placeins}
\usepackage{float}
\usepackage{multirow}
\title{MASS-RAG: Multi-Agent Synthesis Retrieval-Augmented Generation}

\author{
Xingchen Xiao$^{1}$,
Heyan Huang$^{1}$\thanks{Corresponding author.},
Runheng Liu$^{1}$,
Jincheng Xie$^{2}$ \\
$^{1}$School of Computer Science and Technology, Beijing Institute of Technology \\
$^{2}$Department of Mathematical Sciences, Tsinghua University \\
\texttt{\{xcxiao,hhy63,rhliu\}@bit.edu.cn} \\
\texttt{xiejc22@mails.tsinghua.edu.cn}
}
\begin{document}
\maketitle

\begin{abstract}
Large language models (LLMs) are widely used in retrieval-augmented generation
(RAG) to incorporate external knowledge at inference time.
However, when retrieved contexts are noisy, incomplete, or heterogeneous,
a single generation process often struggles to reconcile evidence effectively.
We propose \textbf{MASS-RAG}, a multi-agent synthesis approach to
retrieval-augmented generation that structures evidence processing into
multiple role-specialized agents.
MASS-RAG applies distinct agents for evidence summarization, evidence extraction,
and reasoning over retrieved documents, and combines their outputs through a
dedicated synthesis stage to produce the final answer.
This design exposes multiple intermediate evidence views, allowing the model to
compare and integrate complementary information before answer generation.
Experiments on four benchmarks show that MASS-RAG consistently improves
performance over strong RAG baselines, particularly in settings where relevant
evidence is distributed across retrieved contexts.
\end{abstract}

\begin{figure}[t]
\centering
\includegraphics[
  width=\columnwidth,
  trim=0 0 0 0,
  clip
]{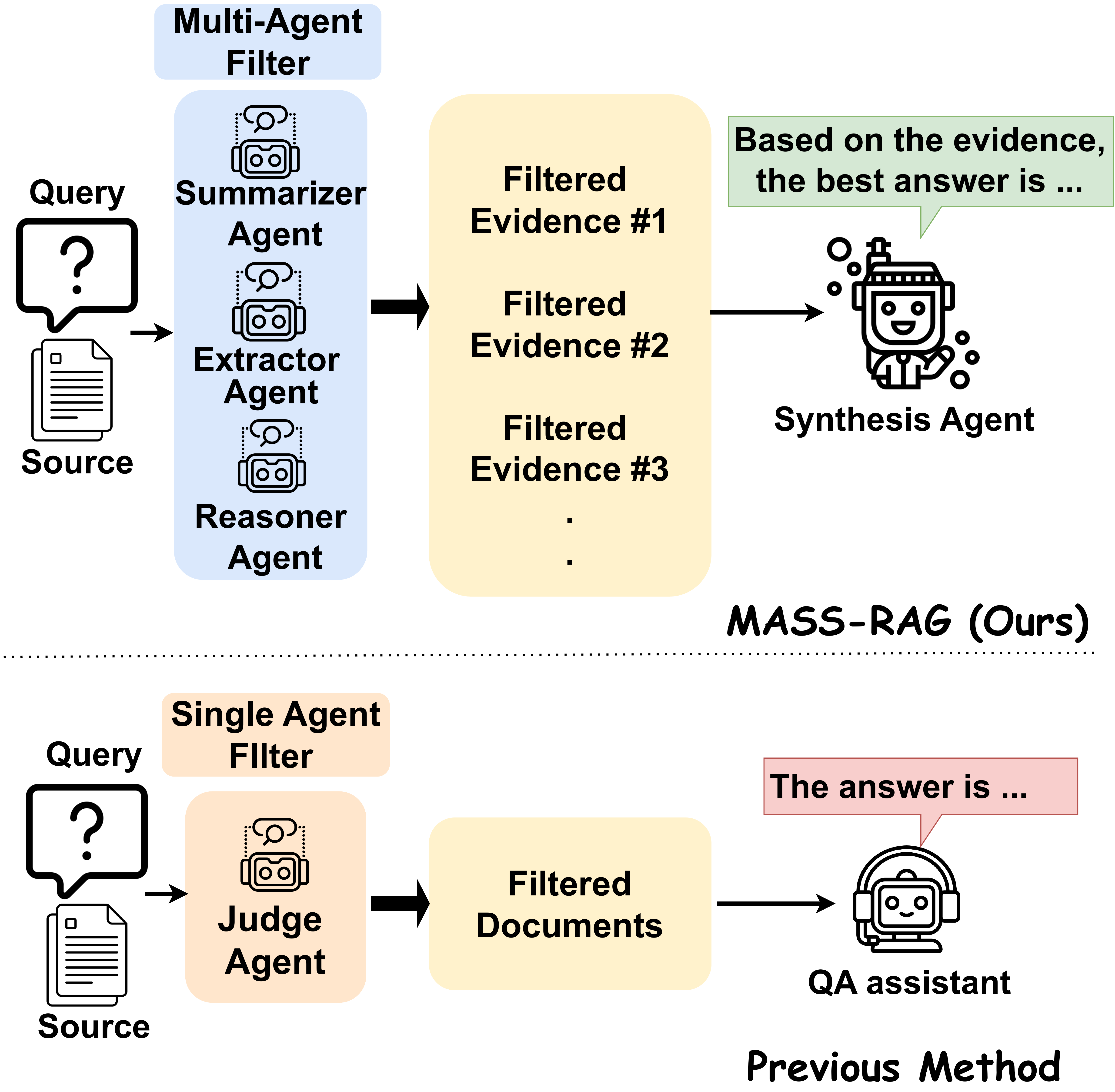}
\caption{A conceptual illustration of the key idea behind MASS-RAG, contrasting the multi-agent evidence filtering and synthesis with the previous single-filter agent method.}
\label{fig:mass_show}
\end{figure}
\section{Introduction}

Large Language Models (LLMs) have demonstrated remarkable scalability and emergent capabilities \citep{brown2020languagemodelsfewshotlearners,wei2022emergentabilitieslargelanguage,touvron2023llama2openfoundation,openai2024gpt4technicalreport}.
However, they remain susceptible to generating unreliable or hallucinatory outputs due to limited factual grounding and static knowledge. To address these limitations, Retrieval-Augmented Generation (RAG) has emerged as an effective paradigm to enhance factual reliability and broaden the knowledge scope of LLMs \citep{huang2023surveyhallucinationlargelanguage,xu2024hallucinationinevitableinnatelimitation,asai2024reliable}.
Updating the parametric knowledge of LLMs through fine-tuning or continual pre-training is both computationally intensive and costly \citep{meng2023locating}, making RAG integration a practical alternative for maintaining up-to-date knowledge at a lower cost \citep{balaguer2024ragvsfinetuningpipelines}.

Previous efforts on the RAG system have explored either pre-training models from scratch\citep{borgeaud2022improving} or continually pre-training existing models to incorporate external knowledge \citep{izacard2022atlas,wang2023instructretro}.
However, such approaches are computationally expensive and resource-intensive, which limits their scalability in real-world RAG applications\citep{gao2024retrievalaugmented}.
As a result, recent RAG research has largely focused on combining retrievers for document selection with large language models (LLMs) for generation\citep{ram2023incontext,lin2023radit,shi2023replug}.
By leveraging external non-parametric knowledge sources, these systems enable knowledge updates without retraining the underlying LLMs. Nevertheless, LLMs remain susceptible to irrelevant or redundant retrieved information\citep{liu2023lostmiddlelanguagemodels,cho2023improvingzeroshotreaderreducing}, which can degrade generation quality.
Consequently, a key challenge is to ensure that RAG systems produce robust and factually accurate outputs when retrieved contexts are noisy or incomplete due to the inherent nature of the retrieval models \citep{gao2024retrievalaugmented,asai2024reliable}. While recent multi-agent RAG framework proposed by \citet{chang2024mainragmultiagentfilteringretrievalaugmented} introduce agent-based context filtering, they typically rely on a single judge agent operating from a monolithic perspective, which limits the system’s ability to capture complementary or heterogeneous forms of factually relevant evidence.

To address this challenge, we introduce \textbf{MASS-RAG}, a Multi-Agent Synthesis framework for retrieval-augmented generation. It operates in a training-free manner and employs multiple agents with specialized roles for context filtering, response generation, and answer synthesis. \textbf{MASS-RAG} integrates multiple specialized agents to capture complementary evidence in the retrieved context, thereby improving robustness and factual accuracy through structured multi-agent evidence filtering and synthesis.

\begin{figure*}[thbp!]
    \centering
    \includegraphics[width=\linewidth]{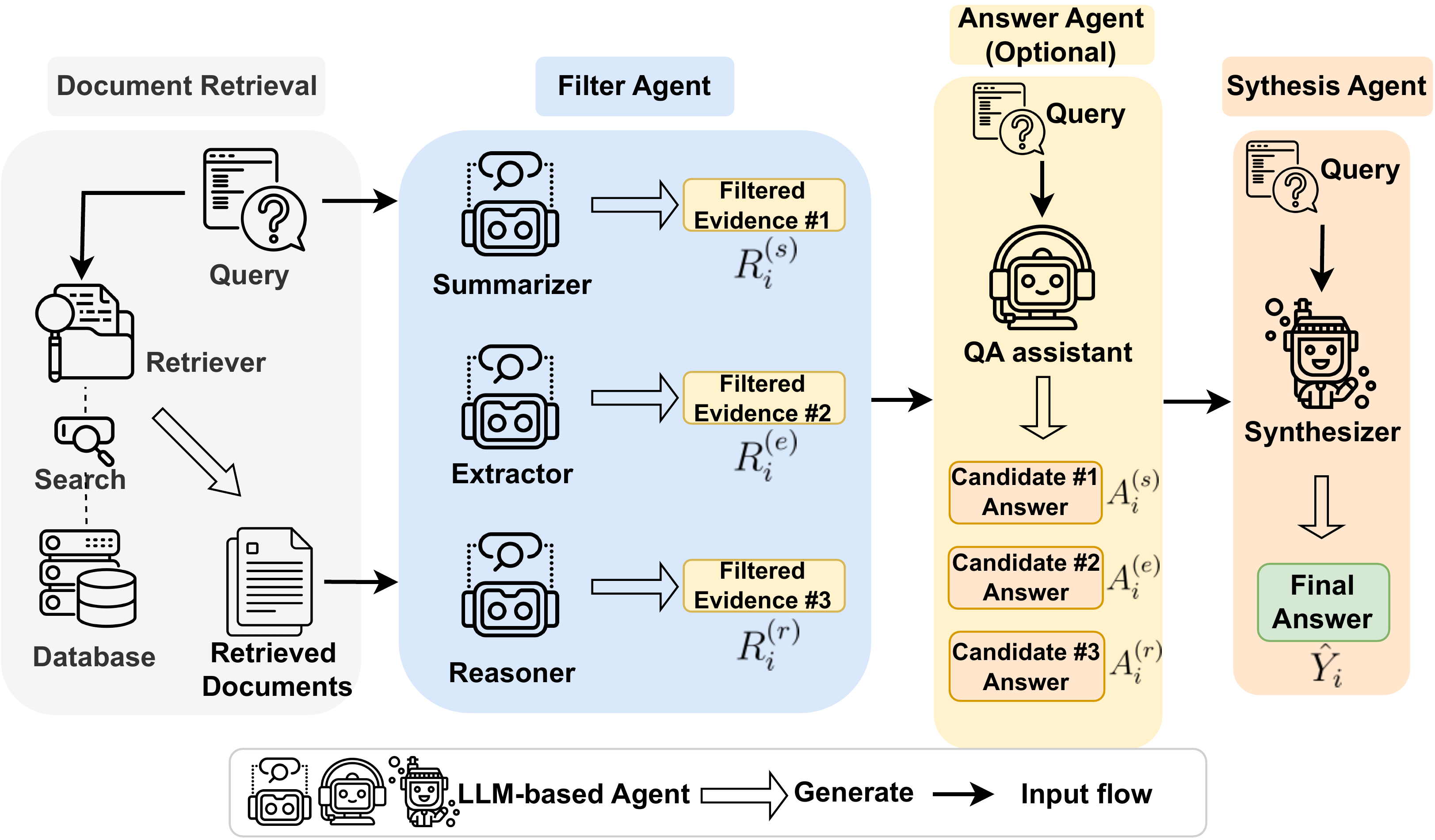}
    \caption{The overall architecture of MASS-RAG, illustrating multi-agent filtering and answer synthesis when the Answer Agent is enabled.}
    \label{fig:main_frame}
\end{figure*}

Our contributions are as follows:
\begin{itemize}
    \item \textbf{Multi-Agent Evidence Filtering:} We introduce a set of specialized LLM-based filter agents that process retrieved documents from complementary perspectives, distilling denoised and query-relevant evidence in a training-free manner.
    \item \textbf{Multi-Agent Answering and Synthesis:} We propose a synthesis mechanism that integrates structured outputs from multiple filter agents, optionally via intermediate candidate answers, to produce a unified final prediction. This design enables explicit comparison across heterogeneous evidence views and supports task-adaptive configurations. 
    \item \textbf{Empirical Validation:} We evaluate MASS-RAG across multiple benchmarks and analysis settings, demonstrating consistent improvements over strong RAG baselines, especially in scenarios requiring the aggregation of complementary evidence.
\end{itemize}

\section{Related Work}

\paragraph{Evidence Refinement in RAG}
% Various methods have been proposed to optimize input context in RAG. One of the most intuitive approaches involves reranking retrieved documents \cite{glass2022re2gretrievererankgenerate}, a technique that has demonstrated success in numerous subsequent studies \cite{ram2023incontext,asai2023selfraglearningretrievegenerate,hwang2024dslrdocumentrefinementsentencelevel,xu2024activeragrevealingtreasuresknowledge}. Other approaches focus on refining the context through techniques such as rewriting the query \cite{ma2023queryrewritingretrievalaugmentedlarge, chan2024rqraglearningrefinequeries}, filtering the context \cite{wang2023learningfiltercontextretrievalaugmented, yang2023prcafittingblackboxlarge}, or summarizing (compressing) relevant content for better generation \cite{xu2023recompimprovingretrievalaugmentedlms, hwang2024dslrdocumentrefinementsentencelevel}. Drawing on prior research, \textbf{MASS-RAG} employs three specialized agents to denoise the retrieved documents, preserving only the essential descriptive and relevant information from different perspectives.

A variety of methods have been proposed to optimize the input context for
retrieval-augmented generation (RAG).
One intuitive line of work focuses on reranking retrieved documents
\cite{glass2022re2gretrievererankgenerate}, which has been widely adopted and
extended in subsequent studies
\cite{ram2023incontext,asai2023selfraglearningretrievegenerate,
hwang2024dslrdocumentrefinementsentencelevel,xu2024activeragrevealingtreasuresknowledge}.
Other approaches aim to refine the input context through query rewriting
\cite{ma2023queryrewritingretrievalaugmentedlarge,chan2024rqraglearningrefinequeries},
context filtering
\cite{wang2023learningfiltercontextretrievalaugmented,yang2023prcafittingblackboxlarge},
or content compression and summarization
\cite{xu2023recompimprovingretrievalaugmentedlms,hwang2024dslrdocumentrefinementsentencelevel}.
These methods primarily operate by improving the relevance or compactness of the
retrieved context prior to generation.
Our work follows the line of training-free methods but differs in that we introduce \textbf{Multi-Agent Synthesis (MASS)}, \textbf{MASS-RAG} adopts a multi-perspective filtering approach,
employing multiple specialized agents to distill complementary and denoised
evidence from retrieved documents while preserving query-relevant information.

% \paragraph{Multi-Agent RAG} Recent works have explored the use of multiple agents to enhance retrieval-augmented generation (RAG). Instead of relying on a single retriever–generator pipeline, multi-agent RAG frameworks distribute the various processes among several specialized agents, 
% each focusing on distinct aspects such as query reformulation \cite{chen2025improvingretrievalaugmentedgenerationmultiagent}, query decompose \cite{nguyen2025maragmultiagentretrievalaugmentedgeneration} and document filtering \cite{chang2024mainragmultiagentfilteringretrievalaugmented}. 
% These methods have shown improved robustness and performance compared to traditional single-pipeline RAG. 
% Our work follows the line of training-free methods but differs in that we introduce a \textbf{Multi-Agent Synthesis Strategy (MASS)}, where different agents filter retrieved documents from distinct perspectives and generate complementary denoised responses, which are subsequently synthesized into a robust final response (answer synthesis).

\paragraph{Multi-Agent RAG}
Recent studies have explored the use of multiple agents to enhance
retrieval-augmented generation.
Instead of relying on a single retriever–generator pipeline, multi-agent RAG
frameworks distribute different sub-processes among specialized agents, such as
query reformulation \cite{chen2025improvingretrievalaugmentedgenerationmultiagent},
query decomposition \cite{nguyen2025maragmultiagentretrievalaugmentedgeneration},
and document filtering \cite{chang2024mainragmultiagentfilteringretrievalaugmented}.
These approaches have demonstrated improved robustness and performance over
traditional single-pipeline RAG systems.
Our work follows the training-free setting of prior multi-agent RAG methods, but
differs in its focus on evidence-centric synthesis.
Specifically, \textbf{MASS-RAG} introduces a multi-agent synthesis framework in
which distinct agents filter retrieved documents from complementary perspectives
and generate denoised evidence representations that are subsequently integrated
to produce a robust final answer.
\section{Multi-Agent Synthesis RAG (MASS-RAG)}
\subsection{Definition of LLM Agents in MASS-RAG}
\label{sec:3.1}

\textbf{MASS-RAG} decomposes retrieval-augmented generation into three stages:
evidence distillation, candidate answer generation, and final answer synthesis.
Accordingly, it consists of several specialized LLM agents, including a Summarizer, an Extractor, a Reasoner, a Synthesis Agent, and an optional Answer Agent. Among them, the Summarizer, Extractor, and Reasoner serve as \emph{filter agents}
that distill query-relevant evidence from retrieved documents, each emphasizing a distinct perspective on evidence processing.

\paragraph{Summarizer Agent}
Prior work has shown that compressing retrieved documents can significantly
improve RAG performance
\cite{xu2023recompimprovingretrievalaugmentedlms,hwang2024dslrdocumentrefinementsentencelevel}.
Building on this insight, the Summarizer agent condenses the retrieved documents
into a concise representation that preserves semantic consistency while
highlighting salient, query-relevant information.
Formally, the summarization process is defined as:
\begin{equation}
    R_i^{(s)} = \mathcal{A}_{\text{sum}}(q_i, D),
\end{equation}
where \( R_i^{(s)} \) denotes the summary-based filtered response.

\paragraph{Extractor Agent}
The Extractor agent identifies fine-grained factual spans or sentences that
explicitly support the answer to \( q_i \) without paraphrasing.
Unlike the Summarizer, this agent strictly operates in an extractive manner,
copying evidence verbatim from the retrieved documents, motivated by previous work on extractive QA\cite{chen2017readingwikipediaansweropendomain,lewis2021retrievalaugmentedgenerationknowledgeintensivenlp}.
It is particularly suited for questions whose answers can be directly grounded
in explicit textual evidence.
This process is expressed as:
\begin{equation}
    R_i^{(e)} = \mathcal{A}_{\text{ext}}(q_i, D),
\end{equation}
where \( R_i^{(e)} \) consists of explicit evidence fragments aligned with the
query semantics.

\paragraph{Reasoner Agent}
The Reasoner agent performs reasoning over the retrieved content to infer
implicit connections or cross-document evidence that may not be explicitly
stated in any single sentence, as motivated by prior work on reasoning and
relevance judgment in retrieval-augmented generation \cite{asai2023selfraglearningretrievegenerate, chang2024mainragmultiagentfilteringretrievalaugmented}.
The Reasoner Agent leverages the backbone language model to analyze the retrieved documents and explicitly articulate how the evidence supports the query, as well as the inference steps required to arrive at an answer.
Formally, it is defined as:
\begin{equation}
    R_i^{(r)} = \mathcal{A}_{\text{rea}}(q_i, D),
\end{equation}
where \( R_i^{(r)} \) represents a reasoning-oriented filtered response that
connects and interprets evidence across retrieved documents.

Notably, MASS-RAG instantiates multiple agents with explicitly differentiated roles and output constraints. Each agent imposes distinct output constraints that explicitly structure how retrieved documents are processed prior to answer generation. Unlike prior approaches, the Reasoner in MASS-RAG produces an explicit intermediate evidence representation that is decoupled from the final answer.

\paragraph{Answer Agent.}
Given each filtered response, the Answer Agent independently generates a
candidate answer. Specifically, for the outputs from the Summarizer,
Extractor, and Reasoner, the Answer Agent computes:
\begin{equation}
    A_i^{(j)} = \mathcal{A}_{\text{ans}}(q_i, R_i^{(j)}), \quad j \in \{s, e, r\},
\end{equation}
where \( A_i^{(s)} \), \( A_i^{(e)} \), and \( A_i^{(r)} \) denote candidate
answers derived from summary-, extraction-, and reasoning-based evidence,
respectively.

Generating candidate answers independently encourages the model to commit
to a concrete answer under each evidence view. This design is most
effective for factoid-style question answering, where candidate answers
carry rich semantic content and different evidence views may yield
complementary or competing hypotheses. In contrast, for multiple-choice
tasks with symbolic option labels (e.g., A/B/C/D), intermediate candidate
answers provide limited additional signal. For long-form QA, the candidate-answer step can still be beneficial, though its effectiveness may be constrained by a tendency to focus on a single dominant response. Accordingly, the Answer Agent can be optionally enabled depending on the task characteristics.

\paragraph{Synthesis Agent.}
The candidate answers
\( \{A_i^{(s)}, A_i^{(e)}, A_i^{(r)}\} \) may overlap while differing in factual
details or reasoning paths.
When the Answer Agent is enabled, a \textbf{Synthesis Agent} integrates these
intermediate answer hypotheses and produces the final output as:
\begin{equation}
    \hat{Y}_i = \mathcal{A}_{\text{syn}}
    \bigl(q_i, \{A_i^{(s)}, A_i^{(e)}, A_i^{(r)}\}\bigr).
\end{equation}

When the Answer Agent is disabled, the Synthesis Agent directly operates on the
filtered evidence responses generated by the filter agents:
\begin{equation}
    \hat{Y}_i = \mathcal{A}_{\text{syn}}
    \bigl(q_i, \{R_i^{(s)}, R_i^{(e)}, R_i^{(r)}\}\bigr).
\end{equation}

The Synthesis Agent is designed to explicitly compare and reconcile structured evidence or answer representations produced by heterogeneous agents, rather than to simply increase model capacity through additional generation steps. Across both settings, this structured synthesis produces a coherent final answer with improved factual accuracy, completeness, and robustness.

We provide detailed role specifications for all agents in
Appendix~\ref{app:agent_roles}.

\begin{table*}[t]
\centering
\small
\resizebox{0.95\textwidth}{!}{
\begin{tabular}{lcccccc}
\toprule
& \multicolumn{1}{c}{TriviaQA}
& \multicolumn{1}{c}{PopQA } 
& \multicolumn{1}{c}{ARC-C} 
& \multicolumn{3}{c}{ASQA} \\
\cmidrule(lr){5-7}
&  (acc) &  (acc) & (acc) & (em) & (rg) & (mau) \\
\midrule
\multicolumn{7}{l}{\textit{Baselines without retrieval}} \\
Llama2$_{7\text{B}}^{*}$ & 30.5 & 14.7 & 21.8 & 7.9 & 15.3 & 19.0 \\
Llama2$_{13\text{B}}^{*}$ & 38.5 & 14.7 & 29.4 & 7.2 & 12.4 & 16.0 \\
Mistral$_{7\text{B}}^{*}$ & 54.8 & 26.2 & 55.5 & 11.2 & 18.1 & 27.6 \\
Llama3$_{8\text{B}}^{*}$ & 68.4 & 29.2 & 58.8 & 19.4 & 30.3 & 54.3 \\
\midrule
\multicolumn{7}{l}{\textit{Baselines with retrieval (training-based)}} \\
Llama2-FT$_{7\text{B}}^{*}$ & 57.3 & 48.7 & 65.8 & 31.0 & 35.8 & 51.2 \\
Self-RAG$_{7\text{B}}^{*}$ & 66.4 & 54.9 & 67.3 & 30.0 & 35.7 & \textbf{74.3} \\
\midrule
\multicolumn{7}{l}{\textit{Baselines with retrieval (training-free)}} \\
Llama2$_{7\text{B}}^{*}$ & 68.9 & 50.9 & 51.0 & 16.2 & 23.4 & 33.1 \\
Llama2$_{13\text{B}}^{*}$ & 47.0 & 45.7 & 26.0 & 16.3 & 20.5 & 24.7 \\
Mistral$_{7\text{B}}^{*}$ & 69.4 & 55.5 & 57.1 & 32.4 & 34.8 & 54.3 \\
Llama3$_{8\text{B}}^{*}$ & 73.1 & 61.8 & 55.6 & 37.1 & 36.5 & 63.0 \\
Qwen3$_{8\text{B}}$ & 72 & 62.1 & \underline{86.4} & 46.1 & 35.7 & 16.1 \\
MAIN-RAG-Mistral$_{7\text{B}}^{*}$ & 71.0 & 58.9 & 58.9 & 35.7 & 36.2 & 60.0  \\
MAIN-RAG-Llama3$_{8\text{B}}^{*}$ & \underline{74.1} & 64.0 & 61.9 & 39.2 & \textbf{42.0} & \underline{70.6} \\

\midrule[0.4pt]
\textbf{MASS-RAG-Llama2$_{7\text{B}}$} & 68.6 & 57.8 & 72.2 & 36.2 & 33.3 & 20.94\\
\textbf{MASS-RAG-Mistral$_{7\text{B}}$} & 73.7 & 60.9 & 72.4 & 41.6 & 37.1 & 69.8\\
\textbf{MASS-RAG-Llama3$_{8\text{B}}$} & \textbf{76.7} & \textbf{64.2} & 78.7 & \underline{47} & 35.2 & 55.2\\
\textbf{MASS-RAG-Qwen3$_{8\text{B}}$} & \underline{74.1} & 63.9 & \textbf{87.3} & \textbf{47.51} & \underline{37.4} & 53.5\\

\bottomrule
\end{tabular}
}
\caption{Overall experimental results on four tasks. \textbf{Bold numbers} refer to the best performance among baselines
without retrieval and training-free baselines, and underline numbers refer to the second-best performance. $ ^{*}$ indicates concurrent results conducted by recent works or original papers. For the metrics, \textit{acc}, \textit{em}, \textit{rg}, and \textit{mau} denote \textit{accuracy}, \textit{str-em}, \textit{rouge}, and \textit{MAUVE}, respectively. }
\label{tab:main_results}
\end{table*}
\section{Experiment}

We conduct comprehensive experiments and ablation studies to evaluate the performance of \textbf{MASS-RAG} from three perspectives:
\begin{itemize}
    \item \textbf{RQ1:} How does MASS-RAG perform compared to training-based RAG methods and existing training-free RAG approaches, including prior multi-agent systems with single-agent filtering?
    \item \textbf{RQ2:} How do individual filter agents capture factually relevant evidence, and to what extent are their contributions complementary?
    \item \textbf{RQ3:} How does aggregating evidence from multiple filter agents affect the end-to-end performance of the RAG system?
\end{itemize}

In addition to these research questions, we include targeted ablation studies to examine the robustness of MASS-RAG to retrieval depth and to analyze the role of the optional Answer Agent.

\subsection{Datasets}
\label{sec:dataset}
\textbf{MASS-RAG} is evaluated using different models and downstream tasks. We conduct our evaluations under zero-shot setting, where each LLM agent is provided with specific task instructions in the in-context approach \cite{ram2023incontext,chang2024mainragmultiagentfilteringretrievalaugmented}.

\paragraph{Open-Domain Question Answering Tasks (ODQA).} Following previous state-of-the-art (SOTA) work \cite{chang2024mainragmultiagentfilteringretrievalaugmented,asai2023selfraglearningretrievegenerate}, we evaluate MASS-RAG on two ODQA datasets: TriviaQA-unfiltered \cite{joshi2017triviaqalargescaledistantly} and PopQA \cite{mallen2023trustlanguagemodelsinvestigating}. As the testing sets of TriviaQA-unfiltered (open) are not publicly available, to make a fair performance comparison, we use the TriviaQA-unfiltered validation and testing sets provided by existing work \cite{asai2023selfraglearningretrievegenerate}. For PopQA, we use the long-tail subset, consisting of 1,399 rare entity queries whose monthly Wikipedia page views are less than 100. 

\paragraph{Long-form Ambiguous Question Answering.} We use a long-form
QA dataset with 948 samples, ALCE-ASQA \cite{stelmakh2023asqafactoidquestionsmeet,gao2023enablinglargelanguagemodels}, to evaluate \textbf{MASS-RAG} performance in ambiguous questions. Each ambiguous sample in the ASQA dataset may contain multiple pieces of evidence and multiple valid answers, making the task particularly challenging. Following prior work \cite{asai2023selfraglearningretrievegenerate, mallen2023trustlanguagemodelsinvestigating, chang2024mainragmultiagentfilteringretrievalaugmented}, we evaluate ASQA using the official metrics, including string exact match (str-em), output diversity measured by MAUVE, and ROUGE.

\paragraph{Closed-set Task} We test the performance of our proposed method on the ARC-Challenge dataset \cite{clark2018thinksolvedquestionanswering} to assess whether multi-agent context filtering benefits RAG performance on a closed-set reasoning task. We use accuracy as an evaluation metric and report on the test set with 1,172 samples.

\subsection{Baselines}
Following the previous methods \cite{asai2023selfraglearningretrievegenerate,chang2024mainragmultiagentfilteringretrievalaugmented}, we evaluated the performance of the \textbf{MASS-RAG} and baselines on four benchmarks. The main evaluation metric is accuracy which has been widely used \cite{asai2023selfraglearningretrievegenerate,chang2024mainragmultiagentfilteringretrievalaugmented}. 

\paragraph{Baselines without retrievals} 
We compared our method with several publicly available pretrained LLMs, including Llama2$_{7\text{B}}$ \cite{touvron2023llama2openfoundation}, Llama3$_{8\text{B}}$ \cite{grattafiori2024llama3herdmodels}, and Mistral$_{7\text{B}}$ \cite{jiang2023mistral7b}.
\paragraph{Baselines with retrievals} We compared our method with both training-based method and training-free method, which incorporate retrieval either during inference or through the training process. For training-based baselines, we consider Self-RAG \cite{asai2023selfraglearningretrievegenerate} and the Llama2-FT$_{7\text{B}}$ which is the Llama2$_{7\text{B}}$ fine-tuned on the same dataset used by Self-RAG but without the reflection tokens or retrieved passages. For training-free baselines, we used the models we included in the baselines without retrievals and we also introduce the latest model Qwen3$_{8\text{B}}$ as reference \cite{yang2025qwen3technicalreport}. Eventually, we compared our method with the previous SOTA (state-of-the-art) multi-agent RAG method, the \texttt{MAIN-RAG} proposed by \citet{chang2024mainragmultiagentfilteringretrievalaugmented}. We also compared our method with standard RAG baselines which do not require additional training and generally prepend the top retrieved documents to the query as the pre-trained LLM input, using the same retriever as all other baselines we include.

\subsection{Main Results (RQ1)}
Table~\ref{tab:main_results} reports the accuracy results of all methods. MASS-RAG consistently outperforms both training-based and training-free baselines across all evaluated tasks. To ensure a fair comparison with Self-RAG, which is trained on Llama2$_{7\text{B}}$, we evaluate MASS-RAG using the same backbone. Under this setting, MASS-RAG achieves absolute accuracy improvements of up to 3.3\%, 5.3\%, 7.3\%, and 20.7\% on the four selected tasks, respectively.

We further compare MASS-RAG with \texttt{MAIN-RAG} using Llama3$_{8\text{B}}$ as the backbone. As shown in Table~\ref{tab:main_results}, MASS-RAG yields improvements of up to 3.5\%, 0.3\%, 27.1\%, and 19.9\% across the same tasks, demonstrating the effectiveness of the proposed multi-agent synthesis strategy.  We visualize these comparisons in Appendix~\ref{appendix:B}.

\subsection{Experimental Settings}
\textbf{MASS-RAG} is compatible with different pre-trained LLMs as backbone models and operates in a fully training-free manner. We use the retrieved
documents released by \citet{asai2023selfraglearningretrievegenerate} for
all experiments. For each query, up to 20 documents are retrieved and
ranked using the pre-trained Contriever trained on MS MARCO
\cite{izacard2022unsuperviseddenseinformationretrieval}. This setting is also adopted by \texttt{Self-RAG} and \texttt{MAIN-RAG}, enabling
a fair and controlled comparison. All models are decoded using greedy generation, with temperature set to 0 and $top\text{-}p$ set to 1.0, ensuring deterministic and reproducible outputs. The number of top retrieved documents is a hyperparameter in our experiments.
We use the top-5 retrieved documents for Llama2\(_{7\text{B}}\) and the top-10
retrieved documents for other models, following the settings adopted by the
corresponding baselines.

For TriviaQA, PopQA, and ASQA, we instantiate all five agents following the
standard \textbf{MASS-RAG} pipeline shown in Figure~\ref{fig:main_frame} to ensure a consistent experimental setup across QA benchmarks. For ARC-Challenge, we instantiate only four agents and exclude the optional Answer Agent.
This setting adapts to the multiple-choice nature of ARC-Challenge,
where answers are restricted to symbolic option labels (e.g., A, B, C, D)
and thus lack semantic granularity.
The structure of ARC-Challenge does not naturally support intermediate
candidate-answer generation. Accordingly, for ARC-Challenge, the final prediction is directly synthesized by the Synthesis Agent from the filtered responses of the three filter agents.

\newcommand{\modeltag}[1]{\textsubscript{\scriptsize #1}}

\FloatBarrier
\begin{table}[!t]
\centering
\small
\setlength{\tabcolsep}{5pt}  % 单栏下略微收紧列间距
\renewcommand{\arraystretch}{1.15}
\begin{tabularx}{\columnwidth}{l *{4}{>{\centering\arraybackslash}X}}
\toprule
Method & TQA & PopQA & ARC-C & ASQA \\
     & (acc) & (acc) & (acc) & (em) \\
\midrule

\multicolumn{5}{l}{\textit{Retrieved Docs = 5}} \\
\texttt{Self-RAG} & 66.4 & 54.9 & 67.3 & 30.0 \\
MASS-RAG\modeltag{(Llama2\(_{7\text{B}}\))} & 68.6 & 57.8 & 72.2 & 36.2 \\
MASS-RAG\modeltag{(Llama3\(_{8\text{B}}\))} & 75.8 & 61.5 & 79.9 & 45.2 \\
MASS-RAG\modeltag{(Mistral\(_{7\text{B}}\))} & 72.7 & 55.7 & 72.3 & 39.5 \\

\midrule
\multicolumn{5}{l}{\textit{Retrieved Docs = 10}} \\
\texttt{MAIN-RAG}\modeltag{(Llama3\(_{8\text{B}}\))} & 74.1 & 64.0 & 61.9 & 39.2 \\
MASS-RAG\modeltag{(Llama3\(_{8\text{B}}\))} & 76.7 & 64.2 & 78.7 & 47.0 \\
MASS-RAG\modeltag{(Mistral\(_{7\text{B}}\))} & 73.7 & 60.9 & 72.4 & 41.6 \\
\bottomrule
\end{tabularx}
\caption{Performance of MASS-RAG on four downstream tasks with different numbers of retrieved documents.}
\label{tab:agent_ablation_1}
\FloatBarrier
\end{table}

\subsection{Ablation Study and Analysis}

\paragraph{Ablation Study on \# of Retrieved Documents}
Due to limitations in retrieval ranking~\cite{asai2024reliable}, factually relevant evidence may not always appear among the top-ranked documents, making RAG performance sensitive to the number of retrieved documents. To assess the robustness to retrieval depth, we evaluate \textbf{MASS-RAG} under varying numbers of retrieved documents, as shown in Table~\ref{tab:agent_ablation_1}. 

Across the evaluated tasks, \textbf{MASS-RAG} exhibits stable performance trends as the retrieval depth varies, demonstrating robustness to changes in the number of retrieved documents. Notably, when fewer documents are selected as retrieval results, \textbf{MASS-RAG} instantiated with Llama3$_{8\text{B}}$ consistently outperforms \texttt{MAIN-RAG} on TriviaQA, ARC-Challenge, and ASQA, indicating that \textbf{MASS-RAG} remains effective even with limited retrieved
context, where robust evidence filtering and synthesis play a more critical role.
\begin{table}[t]
\centering
\normalsize
\setlength{\tabcolsep}{5pt}
\begin{tabular}{lccc}
\toprule
Dataset & Model & w/ Answer & w/o Answer \\
\midrule
\multirow{2}{*}{TriviaQA} & Mistral & \textbf{77.5} & 76.0 \\
         & Llama3  & \textbf{79.9} & 79.3 \\
\midrule
\multirow{2}{*}{PopQA}    & Mistral & 60.9 & \textbf{61.9} \\
         & Llama3  & \textbf{64.2} & 62.6 \\
\midrule
\multirow{2}{*}{ASQA}     & Mistral & 41.6 & \textbf{43.9} \\
         & Llama3  & 47.0 & \textbf{48.0} \\
\bottomrule
\end{tabular}
\caption{Effect of the Answer Agent across different backbone models.
We report performance with and without the Answer Agent on short-form QA(TriviaQA, PopQA) and long-form QA (ASQA).}
\label{tab:answer_agent_ablation}
\end{table}
\paragraph{Ablation Study on the effect of Answer Agent} 
We analyze the role of the Answer Agent by ablating this component across different backbone models and datasets.
In MASS-RAG, three filter agents (Summarizer, Extractor, and Reasoner) are first applied to distill complementary evidence views, and the Answer Agent then independently generates one candidate answer for each filtered response, resulting in three candidate answers corresponding to the three filter agents.

As shown in Table~\ref{tab:answer_agent_ablation}, incorporating the Answer Agent consistently improves performance on factoid QA benchmarks, including TriviaQA and PopQA, across both Mistral and Llama3 backbones. This suggests that independently generating answer hypotheses from multiple evidence views helps reduce ambiguity and enables effective comparison during the synthesis stage.

In contrast, the Answer Agent yields marginal or no improvement on ASQA. Unlike factoid QA, ASQA requires aggregating a diverse set of factual statements into a comprehensive long-form answer.
In this setting, early commitment to individual candidate answers can reduce the flexibility of integrating information distributed across multiple evidence views.
Consistent with this observation, we disable the Answer Agent in our main experiments on ARC-Challenge, where answers are constrained to symbolic options and do not benefit from intermediate answer hypothesis generation. Overall, these results highlight the task-dependent utility of the Answer Agent and demonstrate the flexibility of MASS-RAG, where this component can be optionally enabled depending on task characteristics.

\paragraph{Construction of the Uniquely Attributable Subset (RQ2)}
To better quantify and analyze the contributions of individual filter agent responses, we process the TriviaQA and PopQA datasets using the outputs of MASS-RAG. Specifically, we identify a subset of questions whose ground-truth evidence is captured by
exactly one of the three filter agents, meaning that each such question is captured by a single type of filter agent. This allows us to quantify how many questions can be answered exclusively based on a particular filter agent response. Through this analysis, we demonstrate that different filter agents provide complementary factual evidence that is uniquely captured by each agent, and that all three agents are indispensable, as each contributes distinct and non-overlapping factual evidence to the overall MASS-RAG system.

For PopQA, following the setup in Section~\ref{sec:dataset}, we start from 1,399 questions and construct a \textbf{Uniquely Attributable Subset} by retaining only questions whose ground-truth evidence is captured by exactly one filter agent, while excluding questions captured by multiple agents or by none. This subset enables focused analysis of questions uniquely handled by each agent. The same procedure is applied to TriviaQA for consistency. This analysis reveals the distinct contribution of each filter agent and highlights the complementarity among their responses. Accordingly, the size of the \textbf{Uniquely Attributable Subset} varies with the backbone model, resulting in 609 (TriviaQA) and 108 (PopQA)
samples for Llama3$_{8\text{B}}$, and 559 (TriviaQA) and 85 (PopQA)
samples for Mistral$_{7\text{B}}$.

\paragraph{Effectiveness of Individual Filter Agents (RQ2)}
\begin{figure}[t]
\centering
\includegraphics[
  width=\columnwidth,
  trim=0 0 0 0,
  clip
]{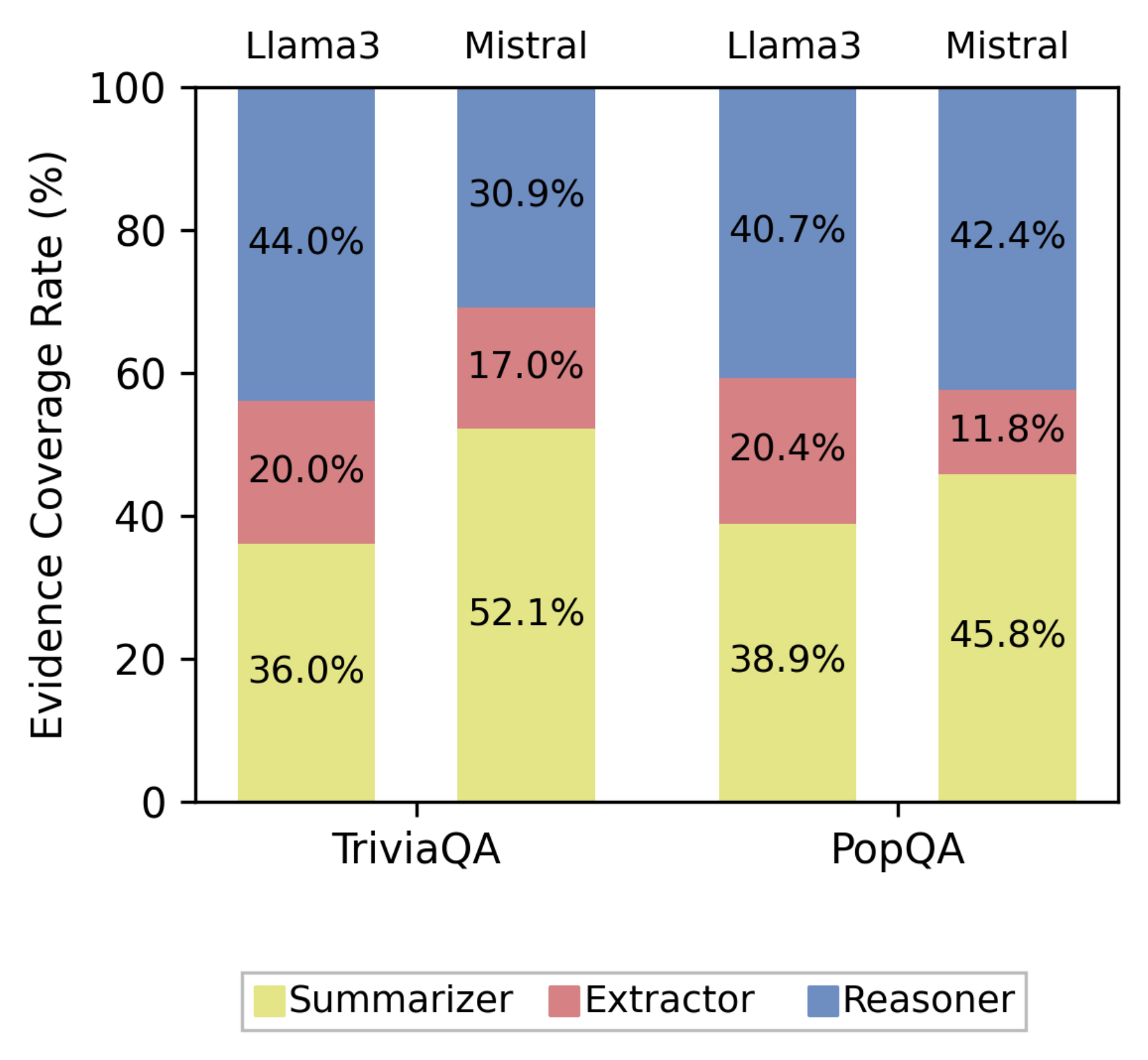}
\caption{Evidence Coverage Rate (ECR) of each filter response on the \textbf{Uniquely Attributable Subset} of TriviaQA and PopQA using Llama3$_{8\text{B}}$ and Mistral$_{7\text{B}}$ with the top-10 retrieved documents.}
\label{fig:ecr_show}
\end{figure}
To evaluate the effectiveness of each type of filter agent response
(\( R_i^{(s)} \), \( R_i^{(e)} \), \( R_i^{(r)} \)) introduced in
Section~\ref{sec:3.1}, we define \textbf{Evidence Coverage Rate (ECR)}, inspired by the Answer Inclusion Rate (AIR) proposed by \citet{ruan2024largelanguagemodelsgood}, as the proportion of questions for which a given filter agent response contains ground-truth evidence. It is defined as:
\begin{equation}
\label{EQ:ECR}
\text{ECR}(A_F) = \frac{1}{|Q|} \sum_{q \in Q}
\mathbf{1}[A_F(q) \text{ relevant}]
\end{equation}
In equation~\ref{EQ:ECR}, $A_F$ denotes a single filter agent or a set of filter agents whose responses are jointly considered, and $\mathbf{1}[\cdot]$ is an indicator function that equals 1 if $A_F(q)$ contains ground-truth evidence for question $q$, and 0 otherwise.

As shown in Figure~\ref{fig:ecr_show}, different filter agents capture
distinct and complementary subsets of factually relevant evidence.
Different agents tend to excel on different subsets of questions, supporting the need for multi-agent filtering, since a single agent alone is insufficient to reliably cover all relevant evidence. Additional case studies are provided to qualitatively demonstrate the complementarity among filter agents (see Appendix~\ref{sec:case_study}).

\paragraph{Impact of Multi-Agent Evidence Aggregation (RQ3)}
Building on the evidence-level robustness observed in RQ2, we investigate whether aggregating evidence from multiple agents enhances the end-to-end performance of the RAG system under noisy retrieval. 

As described in Section~\ref{sec:3.1}, MASS-RAG employs a synthesis agent
$\mathcal{A}_{\text{syn}}$ to produce the final answer by aggregating
candidate answers generated from the responses of the filter agents (\( R_i^{(s)} \), \( R_i^{(e)} \), \( R_i^{(r)} \)).
We evaluate this process using answer accuracy on TriviaQA and PopQA.
\begin{table}[t]
\centering
\normalsize
\begin{tabular}{lcc}
\toprule
Agent  & TriviaQA & PopQA \\
 & (acc) & (acc) \\
\midrule
Summarizer  & 73.7 & 65.7 \\
Extractor   & 68.1 & 61.5 \\
Reasoner    & 72.8 & 63.0 \\
Synthesis   & 76.7 & 64.2 \\
\bottomrule
\end{tabular}
\caption{Accuracy (acc) of each filter response on TriviaQA and PopQA using
Llama3$_{8\text{B}}$ with the top-10 retrieved documents.}
\label{tab:agent_ablation_4}
\end{table} 
\begin{table}[t]
\centering
\normalsize
\begin{tabular}{lcc}
\toprule
Agent  & TriviaQA & PopQA \\
 & (acc) & (acc) \\
\midrule
Summarizer  & 73.8 & 61.2 \\
Extractor   & 66.9 & 57.8 \\
Reasoner    & 71.9 & 62.4 \\
Synthesis    & 73.6 & 60.9 \\
\bottomrule
\end{tabular}
\caption{Accuracy (acc) of each filter response on TriviaQA and PopQA  using Mistral$_{7\text{B}}$ with the top-10 retrieved documents.}
\label{tab:agent_ablation_5}
\end{table} 
As shown in Tables~\ref{tab:agent_ablation_4} and~\ref{tab:agent_ablation_5}, the synthesis agent does not always attain the highest evidence coverage, reflecting its distinct role. 
The summarization agent achieves high coverage, fulfilling its design objective; however, its output serves as an intermediate representation that may include multiple pieces of relevant evidence without resolving their correctness or mutual consistency. Notably, the intermediate responses produced by the filter agents often achieve higher accuracy, indicating that these agents are effective at filtering retrieved documents and exposing answerable content supported by the evidence. In this sense, the filtered responses can be viewed as an approximate upper bound on what the model could potentially answer given the retrieved context. The role of the Synthesis Agent, however, is to produce a single unified answer rather than retain multiple partially correct or ambiguous responses. Whether this upper bound can be reached in the final prediction therefore depends on the base model’s ability to consolidate and resolve the filtered evidence. This observation is consistent with our ablation results, where stronger backbone models are better able to approach this upper bound than weaker ones (e.g. Llama3 vs Mistral).

\section{Conclusion}

We propose MASS-RAG, a multi-agent synthesis framework for retrieval-augmented generation. Experimental results indicate that role-specialized multi-agent filtering captures complementary and factually relevant evidence from retrieved contexts, and that synthesizing these filtered representations leads to more reliable final predictions. Overall, our results suggest that explicitly structuring evidence processing into multi-agent filtering followed by synthesis is a practical design choice for improving the RAG systems.

\section*{Limitations}

Our study is designed to isolate and analyze the effects of multi-agent
evidence filtering and synthesis within a retrieval-augmented generation
framework. To enable controlled comparisons, we adopt fixed retrieval
pipelines and pretrained backbone language models throughout our experiments.
Accordingly, components such as retriever design, reranking strategies,
and decoding-level variations are treated as fixed rather than independent
variables in this work.

MASS-RAG is further evaluated under standard inference-time configurations
without task-specific parameter adaptation. This design choice allows us
to focus on the interaction between agent-level evidence processing and
synthesis, while leaving open how learning-based or adaptive mechanisms
could complement the proposed framework in future studies.

The modular design of MASS-RAG also makes it amenable to future extensions
along orthogonal dimensions, such as retrieval or learning-based adaptation, without altering its core synthesis mechanism.

\section*{Ethical Statement}

This work studies a training-free, multi-agent framework for retrieval-augmented
generation, focusing on how retrieved evidence is filtered and synthesized by
large language models.
The proposed method does not introduce new data sources, supervision signals,
or learning objectives beyond those used in existing RAG systems, and therefore
inherits the ethical considerations associated with the underlying pretrained
models and retrieval corpora.

As with other retrieval-based generation approaches, MASS-RAG may surface
inaccuracies or biases present in retrieved documents or in the backbone
language models.
While the proposed multi-agent filtering and synthesis mechanism aims to improve factual grounding and robustness under noisy retrieval, it does not guarantee the correctness or neutrality of generated outputs.
Accordingly, the system should not be used as a sole source of truth in
high-stakes or safety-critical applications without appropriate human oversight.

We conduct all experiments using publicly available benchmarks and released
retrieval results, and do not involve personal data, user profiling, or
deployment-facing evaluation.
Our goal is to provide a controlled analysis of multi-agent evidence processing rather than to advocate immediate real-world deployment.
\bibliography{custom}

\clearpage
% \onecolumn
\appendix
\section*{Appendix}
% \section*{Appendix}
\label{sec:appendix}

\begin{figure*}[t]
\centering
\includegraphics[
  width=\linewidth,
  trim=0 0 0 0,
  clip
]{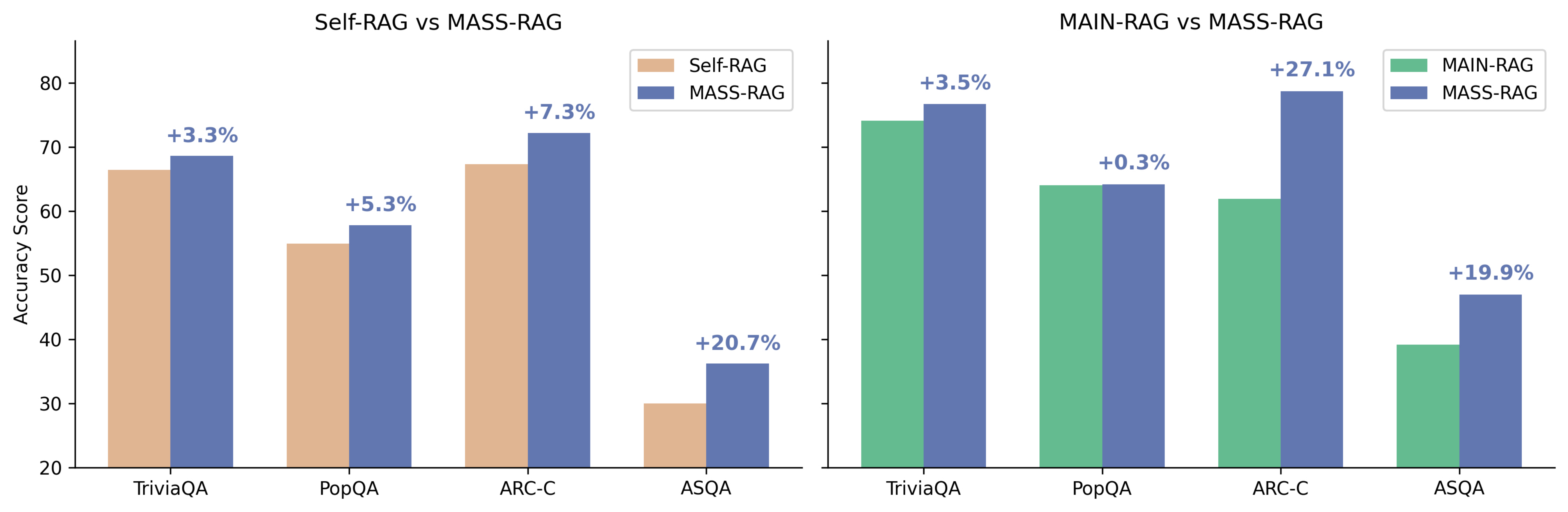}
\caption{Performance comparison of MASS-RAG with state-of-the-art baselines across
four benchmark datasets. The left panel reports results compared with
\textit{Self-RAG} using Llama2$_{7\text{B}}$, while the right panel shows
comparisons with \textit{MAIN-RAG} based on Llama3$_{8\text{B}}$. In both
cases, the backbone models are kept consistent to ensure a fair
comparison.}
\label{fig:mass_comparison}
\end{figure*}

\section{Additional Experimental Setup}
Table~\ref{tab:compute_infra} conclude the computational setup used for
all experiments. Models with 7B or 8B parameters can be executed on a
single 24GB GPU, while models larger than 13B parameters typically
require at least two 24GB GPUs. To ensure a fair comparison with prior
work and to facilitate economical reproducibility, all experiments are
conducted using bfloat16 precision, consistent with the settings adopted
by \textit{Self-RAG}~\cite{asai2023selfraglearningretrievegenerate}.
The total runtime of each experiment may vary depending on system-level
factors such as GPU power limits, runtime load, and driver versions.

\begin{table}[!ht]
\centering
\normalsize
\begin{tabular}{ll}
\toprule
\textbf{Device Attribute} & \textbf{Specification} \\
\midrule
Computing Infrastructure & GPU \\
GPU Model & NVIDIA RTX 4090 \\
Number of GPUs & 4 \\
GPU Memory & 24 GB \\
\bottomrule
\end{tabular}
\caption{Computing Device for the experiments.}
\label{tab:compute_infra}
\end{table}

\section{MASS-RAG Performance Comparison with \textit{SELF-RAG} and \textit{MAIN-RAG}}
\label{appendix:B}
Figure~\ref{fig:mass_comparison} presents a visualized comparison between
\textbf{MASS-RAG} and two representative state-of-the-art baselines,
\texttt{Self-RAG} and \texttt{MAIN-RAG}, across four benchmark datasets.
In the left panel, MASS-RAG is compared with \texttt{Self-RAG} under a shared
Llama2$_{7\text{B}}$ backbone, while the right panel reports comparisons
with \texttt{MAIN-RAG} using Llama3$_{8\text{B}}$ to ensure backbone consistency.
Across all benchmarks, MASS-RAG consistently outperforms both baselines,
with particularly notable gains on reasoning-intensive datasets such as
ARC-C and ASQA. The annotated improvements highlight that MASS-RAG yields
stable performance gains under controlled backbone settings, suggesting
that the observed improvements are attributable to the proposed
multi-agent synthesis framework rather than differences in model variation.

\newpage
\section{Agent Role Specifications}
\label{app:agent_roles}
Each agent in MASS-RAG is instantiated with a lightweight, role-specific instruction that constrains its output objective and format (e.g., extractive, abstractive, or reasoning-oriented), thereby inducing distinct intermediate representations from the same retrieved context.

\paragraph{Summarizer Agent}
The Summarizer is responsible for condensing the retrieved documents into a concise, abstractive summary that highlights query-relevant information. Its output preserves the semantic content of the retrieved context while removing irrelevant or redundant details, and does not attempt to directly answer the query. Even when the retrieved documents do not contain a complete answer, the Summarizer still produces a partial summary of any potentially relevant information.

\paragraph{Extractor Agent}
The Extractor identifies and selects fine-grained textual spans from the retrieved documents that explicitly support answering the query.
It operates in a strictly extractive manner, copying evidence verbatim from the context without paraphrasing or reinterpretation.
When multiple relevant spans exist, the Extractor may return several evidence fragments, reflecting all directly grounded information available in the retrieved documents.

\paragraph{Reasoner Agent}
The Reasoner examines the retrieved documents to identify implicit or cross-document connections that are not explicitly stated in any single sentence. Rather than producing a final answer, it articulates how different pieces of retrieved evidence could be combined or interpreted to support answering the query, including the assumptions or inference steps involved. This agent focuses on reasoning over the retrieved context without introducing external knowledge.

\paragraph{Answer Agent (optional)}
The Answer Agent generates a concise candidate answer based on a single filtered evidence representation produced by one filter agent.
Its role is to translate the provided evidence into a concrete answer hypothesis, without introducing information beyond the given evidence.
When the evidence is partial or incomplete, the Answer Agent produces a best-effort answer that reflects the available information rather than refusing to respond. This agent does not perform cross-evidence aggregation and is applied independently to each filter agent output, serving as an intermediate hypothesis generator rather than the final decision module.

\paragraph{Synthesis Agent}
The Synthesis Agent produces the final output by consolidating structured inputs from multiple agents.
Depending on the task setting, these inputs may consist of filtered evidence representations or intermediate candidate answers.
The Synthesis Agent performs comparative reasoning across its inputs, integrates complementary information, resolves inconsistencies, and generates a coherent final answer without referring to individual agents explicitly.

These role specifications are intentionally lightweight and transferable across backbone language models.

\section{Case Study Of MASS-RAG}
\label{sec:case_study}
\begin{figure*}[!ht]
\centering
\includegraphics[
  width=\linewidth,
  trim=0 0 0 0,
  clip
]{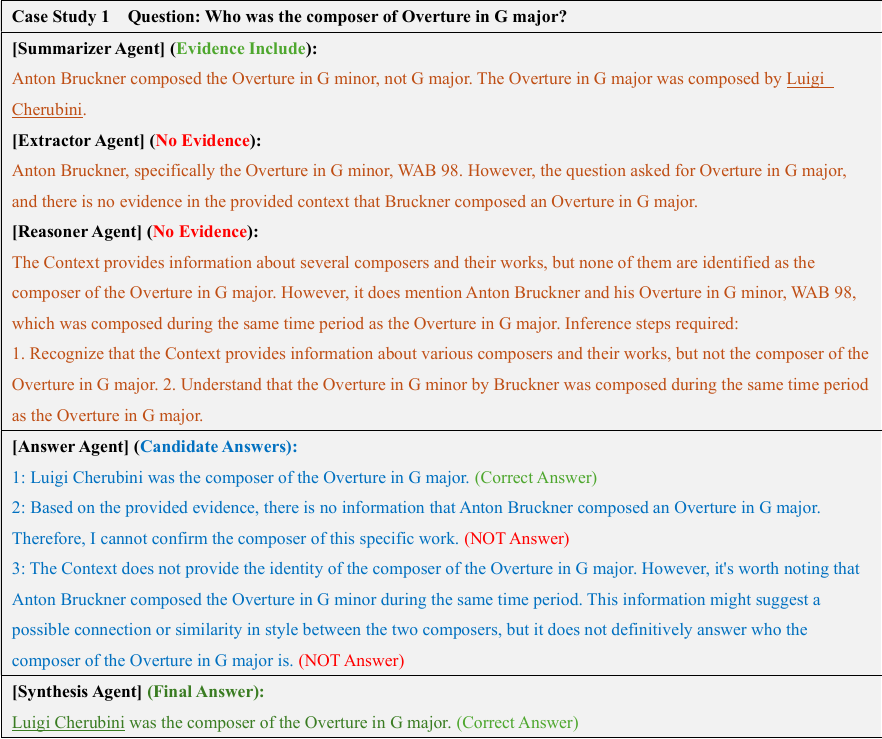}
\caption{MASS-RAG Case Study 1 - Dataset: PopQA, Model: Mistral$_{7\text{B}}$}
\label{fig:Case1}
\end{figure*}

\begin{figure*}[!ht]
\centering
\includegraphics[
  width=\linewidth,
  trim=0 0 0 0,
  clip
]{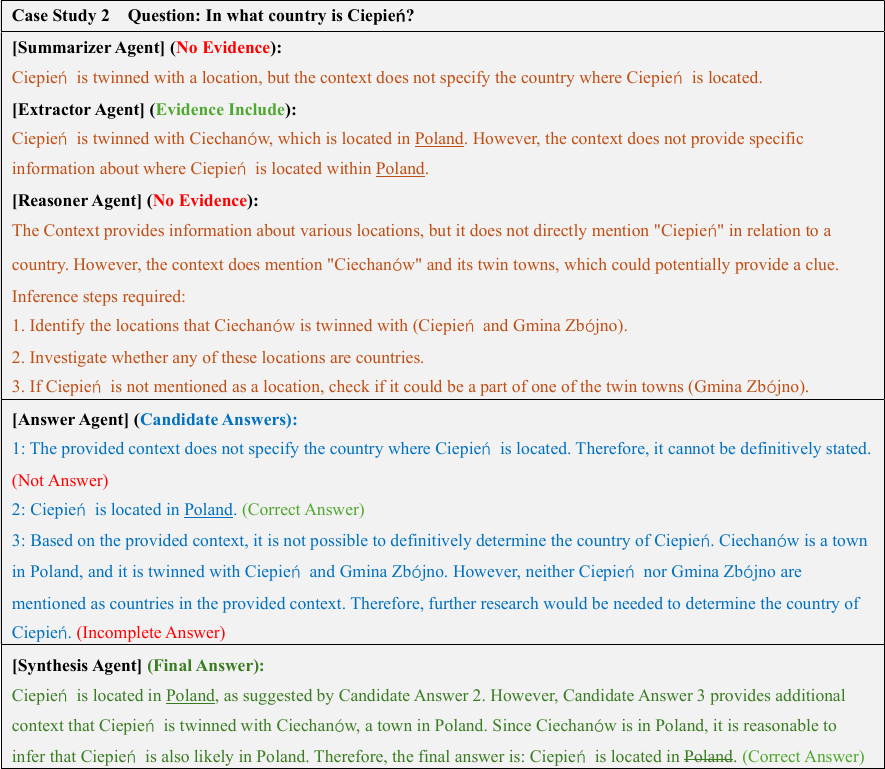}
\caption{MASS-RAG Case Study 2 - Dataset: PopQA, Model: Mistral$_{7\text{B}}$}
\label{fig:Case2}
\end{figure*}

\begin{figure*}[!ht]
\centering
\includegraphics[
  width=\linewidth,
  trim=0 0 0 0,
  clip
]{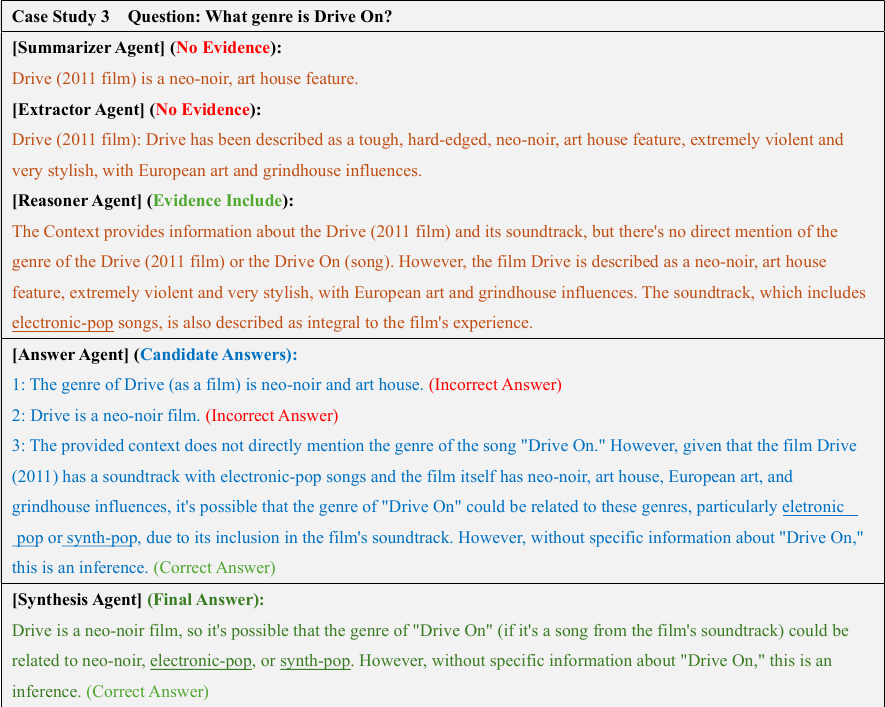}
\caption{MASS-RAG Case Study 3 - Dataset: PopQA, Model: Mistral$_{7\text{B}}$}
\label{fig:Case3}
\end{figure*}

\begin{figure*}[!ht]
\centering
\includegraphics[
  width=\linewidth,
  trim=0 0 0 0,
  clip
]{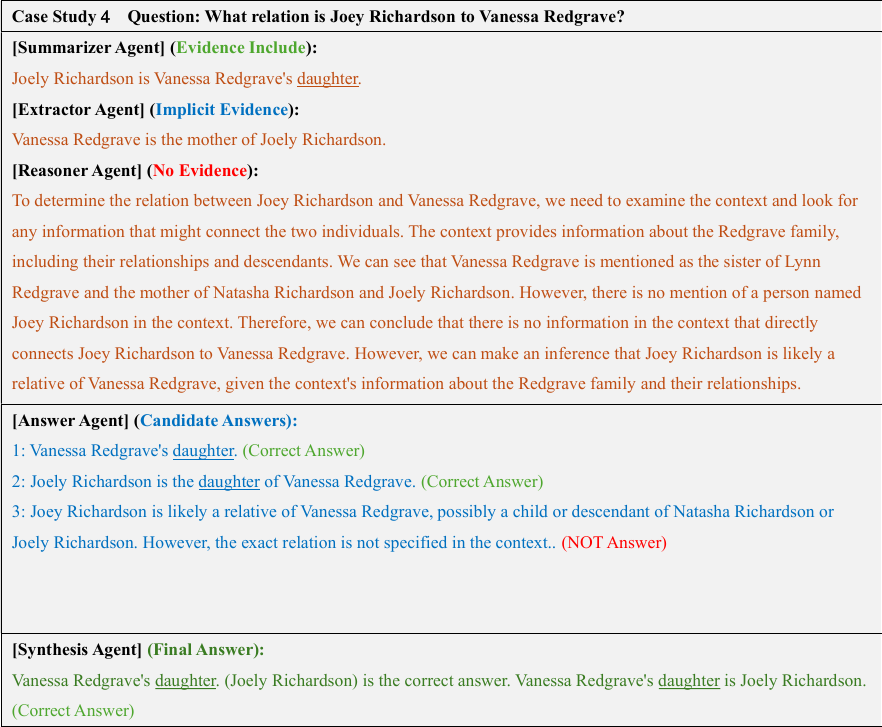}
\caption{MASS-RAG Case Study 4 - Dataset: TriviaQA, Model: Llama3$_{8\text{B}}$}
\label{fig:Case4}
\end{figure*}

\begin{figure*}[!ht]
\centering
\includegraphics[
  width=\linewidth,
  trim=0 0 0 0,
  clip
]{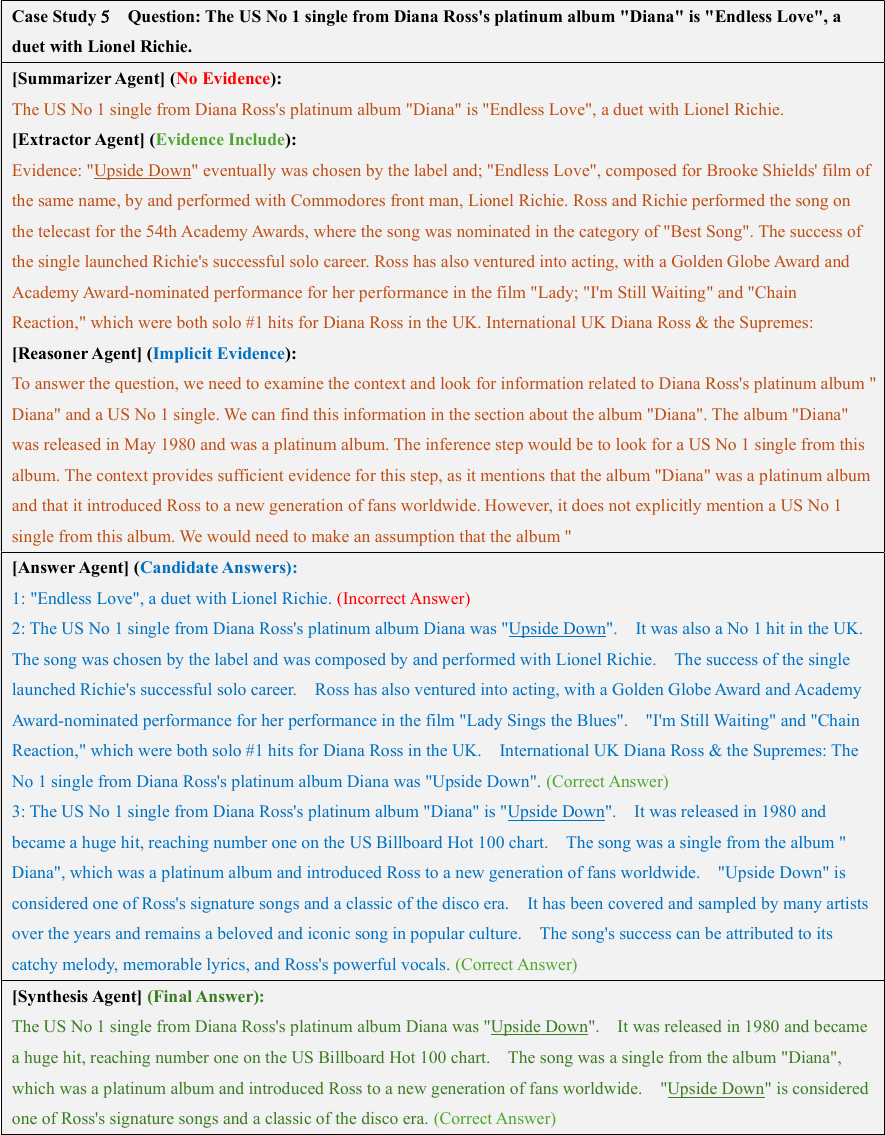}
\caption{MASS-RAG Case Study 5 - Dataset: TriviaQA, Model: Llama3$_{8\text{B}}$}
\label{fig:Case5}
\end{figure*}

\begin{figure*}[!ht]
\centering
\includegraphics[
  width=\linewidth,
  trim=0 0 0 0,
  clip
]{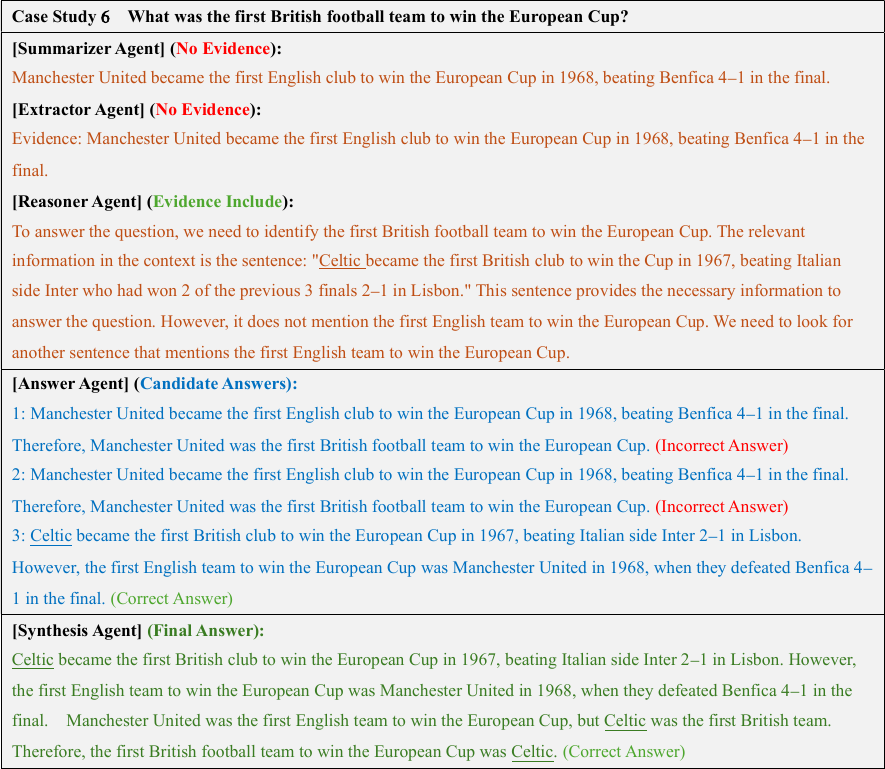}
\caption{MASS-RAG Case Study 6 - Dataset: TriviaQA, Model: Llama3$_{8\text{B}}$}
\label{fig:Case6}
\end{figure*}

We provide additional case studies comprising six examples from Figure~\ref{fig:Case1} to~\ref{fig:Case6}, including
three cases evaluated with Mistral$_{7\text{B}}$ on PopQA and three cases
evaluated with Llama3$_{8\text{B}}$ on TriviaQA. Each case illustrates a
distinct scenario in which only one or two filter agents are able to
capture the relevant factual or implicit evidence. Implicit
evidence refers to evidence that does not explicitly contain the gold
answer, but nonetheless allows the model to infer the correct answer
without requiring additional information. All cases are analyzed used on top-10 retrieved documents setting and the gold answers for each sample are underlined.

\newpage
\section{Runtime Cost Analysis Of MASS-RAG}
In order to analyze the total runtime cost of MASS-RAG, we assume that each agent call costs 100 tokens (Actual settings may vary depends on different task). We use $x$ as the default run time for one agent call. In table \ref{tab:runtime_analysis}, we compare each sample run under different settings.
\label{sec:runtime}
\begin{table}[!ht]
\centering
\normalsize
\begin{tabular}{lll}
\toprule
\textbf{Setting} & \textbf{Run Cost} \\
\midrule
MASS-RAG w/ Answer Agent & 7$x$ \\
MASS-RAG w/o Answer Agent & 4$x$ \\
\bottomrule
\end{tabular}
\caption{Runtime Cost Analysis for different settings.}
\label{tab:runtime_analysis}
\end{table}

\end{document}